\documentclass[journal]{IEEEtran}

\usepackage{subfigure}
\usepackage{palatino,epsfig,latexsym,cite,graphicx,amsmath,amssymb,amsfonts,multirow,booktabs,color,soul}
\usepackage{algorithmic}
\usepackage{float}
\usepackage{threeparttable}
\usepackage{amsmath}
\usepackage[linesnumbered,ruled,vlined]{algorithm2e}
\def\MR2{\multirow{2}[2]{*}}

\definecolor{hl}{rgb}{0.75,0.75,0.75}

\sethlcolor{hl}
\usepackage{makecell}
\usepackage[caption=false,font=footnotesize]{subfig}
\usepackage{xcolor}

\begin{document}

\title{An Evolutionary Algorithm with \\ Probabilistic Annealing for Large-scale Sparse Multi-objective Optimization
\thanks{This work has been submitted to the IEEE for possible publication. Copyright may be transferred without notice, after which this version may no longer be accessible.}
}
\author{
	Shuai Shao,	
	Yuhao Sun,
	Xing Chen,
	Ye Tian*,~\IEEEmembership{Senior Member,~IEEE},
	Guan Wang*,
	and 
	Jin Li*	
	\thanks{Shuai Shao is with the School of Computer Science and Technology, Anhui University, Hefei 230601, China, and also with Sapient Intelligence, Singapore (e-mail: freshshao@gmail.com).}
	\thanks{Yuhao Sun and Xing Chen are with Sapient Intelligence, Singapore (e-mail: intpanty@gmail.com; raincchio@gmail.com).}
	\thanks{Ye Tian is with the School of Computer Science and Technology, Anhui University, Hefei 230601, China (e-mail: field910921@gmail.com).}
	\thanks{Guan Wang and Jin Li are with Sapient Intelligence, Singapore (e-mail: imonenext@gmail.com; electrixoul@outlook.com).}
	\thanks{*Corresponding authors: Ye Tian, Guan Wang, and Jin Li.}
}

\markboth{IEEE Transactions,~Vol.~, No.~, month~year}
{\MakeLowercase{\textit{et al.}}: No title}

\IEEEpubid{0000--0000/00\$00.00~\copyright~0000 IEEE}

\maketitle

\begin{abstract}
Large-scale sparse multi-objective optimization problems (LSMOPs) are prevalent in real-world applications, where optimal solutions typically contain only a few nonzero variables, such as in adversarial attacks, critical node detection, and sparse signal reconstruction. Since the function evaluation of LSMOPs often relies on large-scale datasets involving a large number of decision variables, the search space becomes extremely high-dimensional. The coexistence of sparsity and high dimensionality greatly intensifies the conflict between exploration and exploitation, making it difficult for existing multi-objective evolutionary algorithms (MOEAs) to identify the critical nonzero decision variables within limited function evaluations. To address this challenge, this paper proposes an evolutionary algorithm with probabilistic annealing for large-scale sparse multi-objective optimization. The algorithm is driven by two probability vectors with distinct entropy characteristics: a convergence-oriented probability vector with relatively low entropy ensures stable exploitation, whereas an annealed probability vector with gradually decreasing entropy enables an adaptive transition from global exploration to local refinement. By integrating these complementary search dynamics, the proposed algorithm achieves a dynamic equilibrium between exploration and exploitation. Experimental results on benchmark problems and real-world applications demonstrate that the proposed algorithm outperforms state-of-the-art evolutionary algorithms in terms of both convergence and diversity.
\end{abstract}


\section{Introduction}
\IEEEPARstart{M}{ulti-objective} optimization problems (MOPs) are widely encountered in many scientific and engineering domains, such as structural optimization~\cite{carvalho2024multi}, production scheduling \cite{zhang2025neo}, and financial investment \cite{schnetzer2022evolutionary}. Their core characteristic lies in the simultaneous presence of two or more conflicting objectives, meaning that improving one objective often comes at the expense of others. Therefore, unlike single-objective optimization problems that usually have a unique solution, an MOP corresponds to a set of optimal trade-off solutions among multiple objectives, namely the Pareto optimal solutions. These solutions form a Pareto front in the objective space, depicting the trade-off relationships among different objectives~\cite{tian2025universal,xu2025multi}. The goal of solving an MOP is to obtain a solution set that closely approximates the true Pareto front while maintaining good diversity, thereby providing decision-makers with a wide range of candidate options~\cite{rostamian2025survey,shao2026adaptive}. 

With the increasing complexity of application demands, MOPs often involve a large number of decision variables, thereby giving rise to large-scale MOPs~\cite{si2025efficient,zhong2025indicator}. The resulting high-dimensional search space leads to the so-called ``curse of dimensionality,'' which greatly increases the difficulty of balancing convergence and diversity \cite{mandl2025separable}. Traditional mathematical programming methods, such as linear programming, quadratic programming, and dynamic programming, have been well developed for single-objective optimization~\cite{shir2024multi, tang2025novel}. However, when extended to multi-objective scenarios, they typically rely on scalarization techniques and assumptions of convexity and differentiability of the objective functions, thus showing limited effectiveness in high-dimensional, non-convex, and black-box problems~\cite{lashkar2023multi, ali2025solving}. In contrast, multi-objective evolutionary algorithms (MOEAs), free from such stringent assumptions, can employ flexible operator designs and dimensionality reduction techniques to efficiently explore vast decision spaces, thereby demonstrating stronger adaptability and scalability \cite{yang2025multi, gao2025learnable}.

\IEEEpubidadjcol

In real-world applications, many large-scale MOPs exhibit sparse Pareto optimal solutions, where most decision variables take zero values~\cite{kropp2024improved, shao2025evolutionary}. For instance, in frequent pattern mining~\cite{cao2025pattern, sayari2025robust}, the objective is to identify the most frequent and representative patterns from historical transaction records containing a large number of candidate items. In this process, only a small subset of items is typically selected, resulting in sparse solutions where each decision variable corresponds to whether an item is chosen. Similarly, in the critical node detection~\cite{du2024critical, zhou2024detecting}, the goal is to identify a few nodes that exert the greatest influence on network connectivity or spreading performance. Since only a small fraction of nodes are selected from a large-scale network, this problem also exhibits significant sparsity. Although general large-scale MOEAs have made progress in large-scale multi-objective optimization, their overall efficiency often falls short in addressing such problems due to the high cost of objective function evaluations and the difficulty of directly generating sparse solutions~\cite{shang2025multi, li2025causal}. These problems are collectively referred to as large-scale sparse multi-objective optimization problems (LSMOPs).

In recent years, researchers have proposed a variety of large-scale sparse multi-objective evolutionary algorithms (LSMOEAs) to efficiently solve LSMOPs~\cite{shao2025evolutionary}. These algorithms generally adopt a dual-level encoding scheme, where each solution is represented by a binary vector marking nonzero variables and a real vector refining their values. Since nonzero variables usually account for only a small fraction of all decision variables in solving LSMOPs, accurately identifying them during the optimization process can effectively transform the original large-scale search space into a smaller-scale one, which can then be efficiently solved using existing operators~\cite{ming2024constrained,yin2024adaptive}. Therefore, how to efficiently and accurately identify nonzero variables has become one of the core challenges in large-scale sparse multi-objective optimization. 

To efficiently identify nonzero variables in high-dimensional decision spaces, most LSMOEAs introduce vectors of the same dimension as the decision variables to characterize the propensity of each decision variable to be selected as a nonzero variable~\cite{gu2023quadratic, shao2024deep}. For example, DKCA \cite{li2024dynamic} divides the population’s search space into the original space and a reduced space via a fitness vector, thereby reducing search difficulty. During the process of evolution, DKCA dynamically adjusts the fitness of each variable based on population sparsity, enabling accurate identification of sparse Pareto-optimal solutions. LRMOEA \cite{shao2024evolutionary} archives robust solutions separately and derives a robust guidance vector by analyzing their sparse distribution, which is then used to estimate the probability of each variable being nonzero under uncertain environments. Similarly, IFA \cite{hu2025sparse} introduces an influence factor vector to estimate the likelihood of each decision variable taking a nonzero value and dynamically updates the influence factor vector during the evolutionary process by learning influence factors based on the current population state.

These LSMOEAs leverage vectors to assist genetic operators in efficiently identifying nonzero variables, thereby achieving better results than general large-scale MOEAs~\cite{wang2025spectral,shao2025knowledge}. However, their vectors are often derived from the sparse distribution of the currently obtained locally optimal solutions. As a result, the vectors are limited by local information, hindering further exploration of potential high-quality solutions and constraining the search ability in high-dimensional decision spaces, which easily leads to premature convergence~\cite{shao2025evolutionary}. To overcome these limitations, this paper proposes a probabilistic annealing based multi-population evolutionary algorithm (PAMEA) for large-scale sparse multi-objective optimization. PAMEA employs a dual-entropy probability vector scheme to balance global exploration and local exploitation: high-entropy probability distributions are used to enhance the identification of potential nonzero variables, while low-entropy probability distributions refine the optimization of already identified variables. In addition, PAMEA integrates search processes under different entropy levels through a multi-population cooperation method, which accelerates convergence while maintaining strong exploration ability. Specifically, the main contributions of this paper are as follows:

\begin{enumerate}

	\item A dual-entropy probability vector scheme is proposed. In this scheme, two types of probability vectors are maintained and used collaboratively. The convergence-driven probability vector guides the search at a low entropy level to perform stable exploitation in critical regions, thereby accelerating convergence. In contrast, the annealing-driven probability vector starts with high entropy to enhance global exploration. More importantly, these two probability vectors are coordinated within a multi-population framework, enabling a dynamic balance between exploration and exploitation throughout the evolutionary process. 
	
	\item An annealing-based variable clustering method is developed based on probability vectors. This method evaluates the contribution of each variable group to nonzero variables and assigns an explicit probability to each group to reflect its likelihood of being nonzero, thereby guiding variable flipping. In the early stage of evolution, the contributions are mapped to a narrow probability interval to enhance global exploration. As optimization progresses, the interval is gradually expanded, allowing for fine-grained exploitation of nonzero variables in the later stages.
	
	\item A novel multi-population evolutionary algorithm is constructed by integrating the proposed dual-entropy probability vector scheme and annealing-based variable clustering method. The algorithm leverages probability vectors at different entropy levels to collaboratively identify nonzero variables in high-dimensional decision spaces. To evaluate its effectiveness, extensive empirical studies were conducted on both benchmark problems and real-world applications, comparing the proposed algorithm with state-of-the-art LSMOEAs. Experimental results demonstrate that the proposed algorithm significantly outperforms existing algorithms on most LSMOPs.
	
\end{enumerate}

The remainder of this paper is organized as follows. Section II introduces the key concepts of large-scale sparse multi-objective optimization, provides a systematic classification of existing LSMOEAs, and summarizes their main ideas and limitations, which also motivates this study. Section III details the proposed algorithm. Section IV presents the experimental settings and test platform, followed by a comparative analysis with existing LSMOEAs on a variety of benchmark and real-world LSMOPs. Finally, Section V concludes the paper and outlines future research directions.

\section{Related Work and Motivation}

\subsection{Large-scale Sparse Multi-objective Optimization}

Without loss of generality, an unconstrained multi-objective optimization problem (MOP) can be formulated as
\begin{equation}
	\begin{aligned}
		\min_{\mathbf{z} \in \mathcal{S}} \quad & \mathbf{F}(\mathbf{z}) = \big( f^{(1)}(\mathbf{z}), f^{(2)}(\mathbf{z}), \dots, f^{(m)}(\mathbf{z}) \big)^T \\
		\text{s.t.} \quad & \mathbf{z} = (z_1, z_2, \dots, z_D) \in \mathbb{R}^D
	\end{aligned}
	\label{equ:mop}
\end{equation}
where $\mathbf{z} = (z_1, z_2, \dots, z_D)$ denotes a candidate solution consisting of $D$ decision variables, $F: \mathcal{S} \to \mathbb{R}^m$ is the objective mapping, $\mathcal{S} \subseteq \mathbb{R}^D$ is the decision space, and $\mathbb{R}^m$ is the objective space. In the minimization scenario, given two solutions $\mathbf{z}^a, \mathbf{z}^b \in \mathcal{S}$, if for all $t \in \{1, \dots, m\}$ it holds that $f^{(t)}(\mathbf{z}^a) \leq f^{(t)}(\mathbf{z}^b)$, and at least one inequality is strict, then $\mathbf{z}^a$ is said to Pareto dominate $\mathbf{z}^b$, denoted as $\mathbf{z}^a \prec \mathbf{z}^b$~\cite{ma2023comprehensive,jiang2022evolutionary}. A solution $\mathbf{z}^\star$ that is not dominated by any other solution is called a Pareto-optimal solution. The set of all Pareto optimal solutions is referred to as the Pareto set, and its image in the objective space is known as the Pareto front~\cite{panichella2022improved,kang2024survey}.

Compared with general MOPs, LSMOPs are characterized by an extremely large dimensionality $D$, while in an optimal solution $\mathbf{z}^\star$, only $|\{ i \mid z^\star_i \neq 0 \}| \ll D$ variables are nonzero~\cite{zhou2024evolutionary, shao2025evolutionary}. Although general large-scale MOEAs can be applied to solve LSMOPs, their efficiency is often unsatisfactory in real-world applications due to the high cost of function evaluations~\cite{tian2021evolutionary}. For example, in frequent pattern mining~\cite{cao2025pattern}, evaluating a solution requires traversing the entire transaction database; in critical node detection~\cite{du2024critical}, it requires repeated computations over the whole network; and in neural network training~\cite{keshtkaran2022large}, it requires forward propagation through all training samples. Moreover, most existing large-scale MOEAs are primarily designed for continuous optimization and lack effective strategies to identify nonzero variables, thus making it difficult to generate sparse solution sets. To efficiently approximate sparse Pareto optimal solutions in high-dimensional decision spaces, many customized MOEAs have been developed by exploiting the prior knowledge of solution sparsity in LSMOPs, which will be discussed in the next subsection.

\subsection{Representative LSMOEAs}

SparseEA \cite{tian2020evolutionary} is the first general evolutionary algorithm for solving LSMOPs. Its main contribution lies in representing each solution with a dual-level encoding, which separates the search for nonzero variables from their optimization. Specifically, SparseEA adopts a binary-real representation, where a solution $\mathbf{x}=(x_1,x_2,\dots,x_D)\in\Omega$ is expressed as a combination of a binary vector $\mathbf{xb}=(xb_1,xb_2,\dots,xb_D)\in{0,1}^D$ and a real vector $\mathbf{xr}=(xr_1,xr_2,\dots,xr_D)\in\Omega$, with decision variables computed as $x_i=xb_i\times xr_i,\ i=1,2,\dots,D$. In other words, during the optimization process, the algorithm does not directly store $\mathbf{x}$, but maintains $\mathbf{xb}$ and $\mathbf{xr}$ separately: the binary vector determines which variables take nonzero values, while the real vector further optimizes those nonzero variables. This decoupled representation enables collaborative identification of sparse structures and fine-grained optimization of nonzero variables. It has been adopted by subsequent LSMOEAs and has become a fundamental basis in the field of sparse optimization. Based on the core techniques employed, existing LSMOEAs can be divided into two categories.

The first category constructs vectors to represent the propensity of each variable being nonzero, thereby assisting genetic operators in efficiently generating offspring solutions. LRMOEA~\cite{shao2024evolutionary} stores robust solutions in an archive during evolution, and by analyzing the sparse distribution of archived solutions, it builds robust guidance vectors to estimate the probability of each variable being nonzero under uncertainty, thus guiding the generation of robust offspring. DPREA~\cite{yu2025solving} incorporates prior sparsity information into the initialization of the population and assigns initial scores to each decision variable. During evolution, inspired by reinforcement learning, DPREA dynamically updates the score vector, thereby accelerating the identification of nonzero variables. Similarly, IFA~\cite{hu2025sparse} calculates an influence factor for each decision variable to measure its importance and guide population evolution, with dynamic updates according to iteration progress and solution quality. MOEA-SD~\cite{yang2025sparse} introduces a sparsity detection strategy in the initialization phase, dividing decision variables into nonzero and zero subsets through importance vectors, providing the population with a better starting point. ASD-MOEA~\cite{qiu2025adaptive} dynamically adjusts the fitness vector of decision variables based on iteration progress, flipping more variables in later stages to focus computational resources on optimizing nonzero variables, thereby approaching sparse Pareto-optimal solutions more efficiently.

The second category employs variable clustering or machine learning techniques to reduce the dimensionality of the decision space. For example, DSGEA~\cite{zou2023evolutionary} groups decision variables according to the proportion of nonzero variables and flips multiple variables within each group simultaneously during evolution, significantly accelerating convergence. To further enlarge the flipping granularity, AGSEA~\cite{shao2024deep} partitions decision variables into two subsets of different sizes and flips variables within each group with equal probability during evolution. PM-MOEA \cite{tian2020pattern} applies evolutionary pattern mining to identify candidate sets of maximum and minimum critical variables, thus narrowing the search space of genetic operators. Based on this, TELSO~\cite{qi2024two}enhances population diversity by extracting frequent itemsets across different individuals and introduces a dynamic mutation strategy to avoid premature convergence. EPMEA \cite{qi2024enhancing} designs customized objective functions to assist candidate set mining, where the selection criteria consider not only the number of nonzero variables but also the proportion of nonzero variables in specific dimensions. DKCA \cite{li2024dynamic} identifies critical variables through a variable selection process to construct a reduced decision subspace. During evolution, the algorithm maintains global exploration in the original high-dimensional space while conducting targeted search in the reduced subspace, thereby utilizing computational resources more effectively.

\subsection{Motivation of This Work}

While the aforementioned LSMOEAs have improved the ability to identify nonzero variables to varying extents, their performance still has limitations~\cite{wang2021enhanced, shao2025evolutionary}. Specifically, the first category of  methods typically constructs vectors by exploiting the sparse distribution of the current locally optimal solutions with a fixed entropy level. Therefore, the obtained information is constrained by local solutions, causing the algorithm to lack sustained exploration of potential high-quality solutions and making it prone to local optima in high-dimensional search spaces. The second category of methods reduces the dimensionality of the decision spaces through variable clustering or machine learning techniques. However, these algorithms usually flip all variables in the reduced subspace indiscriminately, failing to fully leverage the probability information of variables taking non-zero values~\cite{zou2023evolutionary, jiang2024improving}. Consequently, they neglect the importance of different variables within the sparse structure, which limits the effective identification of critical variables.

To address the above shortcomings, this paper introduces a dual-entropy probability vector scheme. It simultaneously maintains and collaboratively utilizes two types of probability vectors, namely the convergence-oriented vector and the annealing-driven vector, so that genetic operators can generate offspring solutions under different entropy levels. The convergence-oriented probability vector operates at a lower entropy level and drives the population to perform stable exploitation around already discovered critical regions. The annealing-driven probability vector starts at a high entropy level and enables the population to explore potential high-quality solutions. On this basis, we further propose an annealing-based variable clustering method guided by probability vectors. This method comprehensively evaluates the contribution of each variable within a group to the formation of nonzero variables, and adaptively assigns probabilities to guide the flipping of decision variables. In this way, variables within different groups may be flipped into nonzero states with corresponding probabilities. By combining the dual-entropy probability vectors with the annealing-based variable clustering method, the proposed algorithm can dynamically adjust search behaviors throughout the evolutionary process. This accelerates convergence while maintaining sufficient exploration, thereby improving optimization performance in high-dimensional sparse decision spaces.

\section{The Proposed Algorithm}

\subsection{Procedure of PAMEA}

\begin{algorithm}[!t]
	\caption{The Framework of the Proposed Algorithm}
	\label{alg:main}
	\small
	\begin{algorithmic}[1]
		\REQUIRE $N$ (population size), $FE_{max}$ (maximum number of evaluations), $S$ (number of sampling cycles), $D$ (number of decision variables)
		\ENSURE $P$ (final population)
		\STATE $\mathbf{cpv} \leftarrow$ Generate convergence-driven probability vector using Algorithm~\ref{alg:cpv}\;
		\STATE $P \leftarrow$ Generate initial population using Algorithm~\ref{alg:init};				
		\STATE $FE \leftarrow$ Number of consumed evaluations;
		\WHILE{$FE < FE_{max}$}
        \STATE $\mathbf{apv} \leftarrow$ Generate annealing-driven probability vector using Eq.~\ref{equ:apv}\;
        \STATE $P1\leftarrow$ Select $N$ parents from $P$ via binary tournament selection\;
        \STATE $Q1 \leftarrow$ Generate $N/2$ offspring solutions based on $\mathbf{cpv}$ using Algorithm~\ref{alg:CPVSearch}\;
        \STATE $[Groups, Probs] \leftarrow$ Divide decision variables into groups with different probabilities based on $\mathbf{apv}$ using Algorithm~\ref{alg:VariableClustering}\;       
        \STATE $P2\leftarrow$ Select $N$ parents from $P$ via binary tournament selection\;
        \STATE $Q2 \leftarrow$ Generate $N/2$ offspring solutions based on $Groups$ and $Probs$ using Algorithm~\ref{alg:APVSearch}\; 
		\STATE $P\leftarrow$ Select $N$ solutions from $P\cup Q1 \cup Q2$ via the environmental selection strategy of SPEA2\;		
		\STATE $FE \leftarrow FE + N$;		
		\ENDWHILE
		\RETURN $P$
	\end{algorithmic}
\end{algorithm}

\begin{figure*}[!t]
	\centering
	\subfigure{\includegraphics[width=0.95\linewidth]{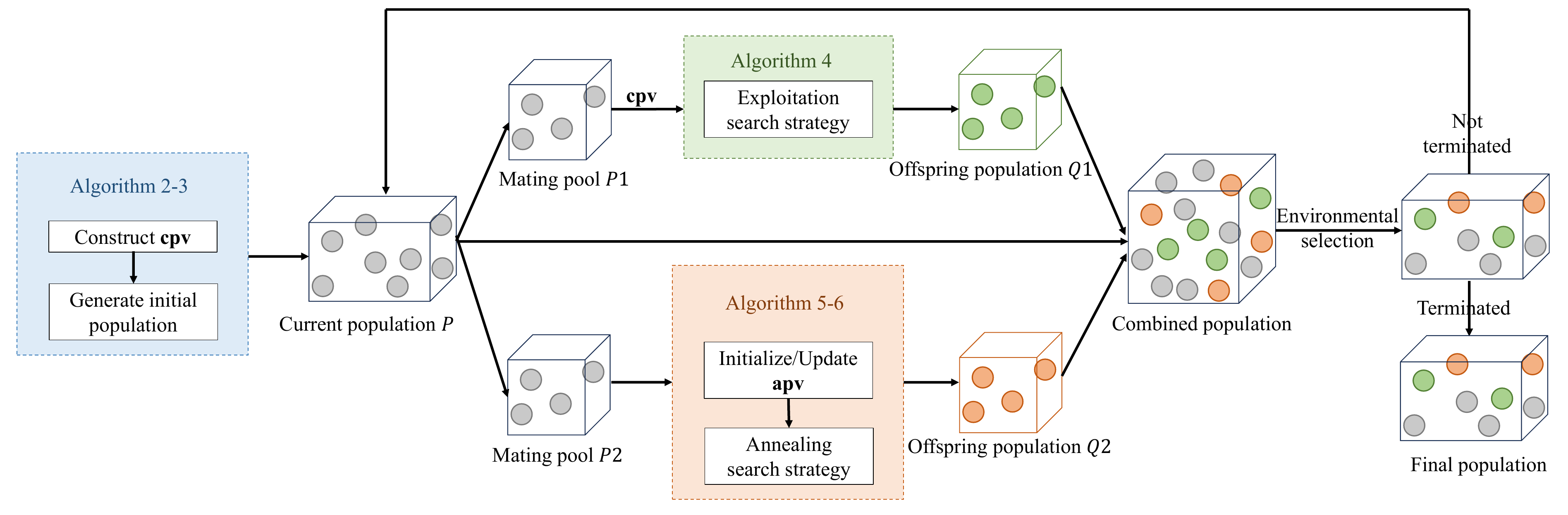}}
	\caption{Illustration of the proposed PAMEA built on the dual-entropy scheme, where an exploitation search strategy and an annealing search strategy are integrated to balance exploitation and exploration for searching sparse Pareto optimal solutions.}
	\label{fig:flowchart}
\end{figure*}

The workflow of the proposed PAMEA is shown in Fig.~\ref{fig:flowchart}, and its framework is given in Algorithm~\ref{alg:main}. It employs two subpopulations that, guided respectively by a convergence-driven probability vector $\mathbf{cpv}$ and an annealing-driven probability vector $\mathbf{apv}$, perform local search and global exploration to approximate sparse Pareto optimal solutions. The $\mathbf{cpv}$ is computed from prior knowledge and, with a low-entropy setting, steers a subpopulation to stably exploit key regions and accelerate convergence. In contrast, the $\mathbf{apv}$ is derived from the population’s sparsity pattern and the current probability intervals; its entropy decays from high to low over the course of evolution, thereby exploring potential high-quality solutions. Built on this dual-entropy scheme, PAMEA consists of two novel components: (i) an exploitation search strategy that performs slight bit flips for local search under the guidance of $\mathbf{cpv}$; and (ii) an annealing search strategy that groups variables using $\mathbf{apv}$, then samples by group probabilities to search for nonzero variables. These two components collaborate to strike a better balance between exploitation and exploration when searching sparse Pareto optimal solutions.

As shown in Algorithm~\ref{alg:main}, $\mathbf{cpv}$ is first constructed via Algorithm~\ref{alg:cpv}, wherein the importance of each decision variable is assessed a priori through statistical sampling. Next, the initial population $P$ is generated using Algorithm~\ref{alg:init}, and $FE$ is set to the total number of function evaluations consumed so far, including those used for $\mathbf{cpv}$ construction and population initialization. In each generation, two complementary search strategies are executed: the first subpopulation adopts the exploitation search strategy, and the second adopts the annealing search strategy. For the exploitation search strategy, the proposed algorithm performs binary tournament selection to obtain the parent set $P_1$ and, under the guidance of $\mathbf{cpv}$, uses Algorithm~\ref{alg:CPVSearch} to produce offspring $Q_1$, thereby reinforcing stable exploitation in key regions. In contrast, the annealing search strategy is more exploratory: Algorithm~\ref{alg:VariableClustering} first partitions the decision variables into several groups according to $\mathbf{apv}$ and assigns each group an explicit probability that reflects its likelihood of containing nonzero variables; guided by this clustering, Algorithm~\ref{alg:APVSearch} generates the second offspring set $Q_2$. Next, the parents and both offspring sets are merged, and the SPEA2 environmental selection mechanism is applied to $P \cup Q_1 \cup Q_2$ to select $N$ individuals as the next generation. After updating $FE$, the process repeats until the maximum number of evaluations $FE_{\max}$ is reached.

To effectively generate sparse solutions, the proposed algorithm adopts a dual-level encoding scheme in which each solution \(\mathbf{x}\) is represented by the Hadamard product of a binary vector \(\mathbf{bin}\) and a real vector \(\mathbf{real}\). The binary vector \(\mathbf{bin}\) controls whether each variable is activated as nonzero (1 indicates a nonzero variable, and 0 indicates a zero variable), while the real vector \(\mathbf{real}\) determines the specific values of the activated variables. In the following three subsections, the three core components of the proposed algorithm will be presented in detail, namely the exploitation search strategy, the annealing-based variable clustering method, and the annealing search strategy.

\subsection{Exploitation Search Strategy}

\begin{algorithm}[!t]
	\caption{CPVCalculate}
	\small
	\label{alg:cpv}
	\begin{algorithmic}[1]
		\REQUIRE $S$ (number of sampling cycles), $D$ (number of decision variables)
		\ENSURE $\mathbf{cpv}$ (convergence-driven probability vector)
		\STATE $C \leftarrow$ Generate $S$ values for each decision variable by the Latin hypercube sampling		
		\STATE $\mathbf{cpv} \leftarrow 1 \times D$ vector of zeros
		\FOR{$s = 1$ to $S$}
		\STATE $Real \leftarrow D \times D$ matrix generated, where each column corresponds to the $i$-th value of the decision variables in $C$
		\STATE $Bin \leftarrow D \times D$ identity matrix
		\STATE $Q \leftarrow$ A population whose $i$-th solution is generated by the $i$-th rows of $Real \odot Bin$
		\STATE $[F_1, F_2,\dots] \leftarrow$ Do non-dominated sorting on $Q$
		\FOR{$i = 1$ to $D$}
		\STATE $k \leftarrow $ The non-dominated front number of the $i$-th solution in $Q$
		\STATE $cpv_i \leftarrow cpv_i + k$
		\ENDFOR		
		\ENDFOR
        \STATE $\mathbf{cpv} \leftarrow 1 - \dfrac{\mathbf{cpv} - \min(\mathbf{cpv})}{\max(\mathbf{cpv}) - \min(\mathbf{cpv})}$
		\RETURN $\mathbf{cpv}$		
	\end{algorithmic}
\end{algorithm}

\begin{algorithm}[!t]
	\caption{Initialization}
	\small
	\label{alg:init}
	\begin{algorithmic}[1]
		\REQUIRE $N$ (population size), $\mathbf{cpv}$ (convergence-driven probability vector), $D$ (number of decision variables)
		\ENSURE $P$ (initial population)
		\STATE $Real \leftarrow$ Uniformly randomly generate the decision variables of $N$ solutions
		\STATE $Bin \leftarrow N \times D$ matrix of zeros
		\FOR{$i = 1$ to $N$}
		\FOR{$j = 1$ to $\lceil rand() \times D \rceil$}
		\STATE $[m, n] \leftarrow$ Randomly select two decision variables
		\IF{$cpv_m > cpv_n$}
		\STATE Set the $m$-th element in the $i$-th row of $Bin$ to 1
		\ELSE
		\STATE Set the $n$-th element in the $i$-th row of $Bin$ to 1
		\ENDIF
		\ENDFOR
		\ENDFOR
		\STATE $P \leftarrow$ A population whose $i$-th solution is generated by the $i$-th rows of $Real \odot Bin$
		\RETURN $P$
	\end{algorithmic}
\end{algorithm}

\begin{algorithm}[!t]
	\caption{ExploitationSearch}
	\small
	\label{alg:CPVSearch}
	\begin{algorithmic}[1]
		\REQUIRE $P^\prime$ (parent solutions), $\mathbf{cpv}$ (convergence-driven probability vector)
		\ENSURE $Q$ (offspring population)		
		\WHILE{$P^\prime\neq\emptyset$}
		\STATE $[\mathbf{p},\mathbf{q}]\leftarrow$ Randomly select two parents from $P^\prime$       	
		\STATE $P^\prime\leftarrow P^\prime\setminus\{\mathbf{p},\mathbf{q}\}$ \\		
		\COMMENT{$\mathbf{bin}^p$ denotes the binary vector of solution $\mathbf{p}$}
		\STATE $index \leftarrow xor(\mathbf{bin}^p, \mathbf{bin}^q)$\;
		\STATE $\mathbf{bin}^o\leftarrow \mathbf{bin}^p$\\
		\COMMENT{Crossover}
		 
		\IF{$rand()<0.5$}            
		\STATE $[m,n] \leftarrow$ Randomly select two bits from $index$\;
        \STATE $k \leftarrow \arg\max_{i \in \{m,n\}} \mathbf{cpv}_i$\;
        \STATE $\mathbf{bin}^o_{k} \leftarrow 1$\;
		\ELSE
		\STATE $[m,n] \leftarrow$ Randomly select two bits from $index$\;
        \STATE $k \leftarrow \arg\min_{i \in \{m,n\}} \mathbf{cpv}_i$\;
        \STATE $\mathbf{bin}^o_{k} \leftarrow 0$\;
		\ENDIF\\		
		\COMMENT{Mutation}
		\IF{$rand()<0.5$}		
		\STATE $[m,n] \leftarrow$ Randomly select two bits from the zero bits in $\mathbf{bin}^o$\;
        \STATE $k \leftarrow \arg\max_{i \in \{m,n\}} \mathbf{cpv}_i$\;
        \STATE $\mathbf{bin}^o_{k} \leftarrow 1$\;
		\ELSE	
		\STATE $[m,n] \leftarrow$ Randomly select two bits from the nonzero bits in $\mathbf{bin}^o$\;
        \STATE $k \leftarrow \arg\min_{i \in \{m,n\}} \mathbf{cpv}_i$\;
        \STATE $\mathbf{bin}^o_{k} \leftarrow 0$\;
		\ENDIF \\
		\COMMENT{Crossover and mutation for real vectors}
		\STATE $\mathbf{real}^o\leftarrow$ Perform simulated binary crossover and polynomial mutation on $\mathbf{\mathbf{real}^p}$ and $\mathbf{real}^q$
		\STATE $\mathbf{o}\leftarrow$ A solution determined by $\mathbf{bin}^o$ and $\mathbf{real}^o$
		\STATE $Q\leftarrow Q\cup\{\mathbf{o}\}$
		\ENDWHILE		
		\RETURN $Q$
	\end{algorithmic}
\end{algorithm}

\begin{algorithm}[!t]
  \caption{VariableClustering}
  \small
  \label{alg:VariableClustering}
  \begin{algorithmic}[1]
    \REQUIRE $P$ (current population), $\mathbf{apv}$ (annealing-driven probability vector), $D$ (number of decision variables)
    \ENSURE $Groups$ (variable groups), $Probs$ (probability assigned to each group)

    \COMMENT{Compute average  sparsity over the population}
    \STATE $\bar{\rho} \leftarrow \dfrac{1}{|P|}\sum_{p\in P}\dfrac{\|\mathbf{bin}^p\|_0}{D}$;

    \COMMENT{Determine target group size}
    \STATE $G_{\text{size}} \leftarrow \max\{1,\ \mathrm{round}(\bar{\rho}\cdot D)\}$;
    \STATE $G_{\text{size}} \leftarrow \min\{G_{\text{size}},\ D\}$;

    \STATE $\pi \leftarrow$ Sort variable indices by $\mathbf{apv}$ in descending order;

    \COMMENT{Partition variables into contiguous groups and assign group probabilities}
    \STATE $K \leftarrow \left\lceil \dfrac{D}{G_{\text{size}}} \right\rceil$;
    \STATE $Groups \leftarrow \emptyset$;
    \STATE $Probs \leftarrow \emptyset$;
    \FOR{$g = 1$ to $K$}
      \STATE $s \leftarrow (g-1)\cdot G_{\text{size}} + 1$;
      \STATE $t \leftarrow \min\{g\cdot G_{\text{size}},\ D\}$;
      \STATE $G_g \leftarrow \{\pi_s,\ \pi_{s+1},\ \dots,\ \pi_t\}$;

      \COMMENT{Group probability equals the mean of $\mathbf{apv}$ over variables in $G_g$}
      \STATE $p_g \leftarrow \dfrac{1}{|G_g|}\sum_{i\in G_g}\mathbf{apv}_i$;

      \STATE $Groups \leftarrow Groups \cup \{G_g\}$;
      \STATE $Probs \leftarrow Probs \cup \{p_g\}$;
    \ENDFOR

    \RETURN $Groups$, $Probs$
  \end{algorithmic}
\end{algorithm}

In large-scale sparse multi-objective optimization, rapidly identifying nonzero variables is crucial for efficiently generating high-quality offspring solutions. To this end, this paper introduces a prior vector $\mathbf{cpv}$ constructed based on sampling statistics to estimate, before the evolution begins, the probability of each variable being activated as nonzero, thereby guiding the local search to accelerate convergence. Algorithm~\ref{alg:cpv} presents the procedure for computing $\mathbf{cpv}$. Specifically, Latin hypercube sampling is first employed to generate $S$ sample values for each decision variable, forming a sampling matrix $C$, while initializing $\mathbf{cpv}$ as a zero vector. In each sampling iteration, a real matrix $Real \in \mathbb{R}^{D\times D}$ and a binary unit matrix $Bin \in {0,1}^{D\times D}$ are constructed. Their Hadamard (element-wise) product $\odot$ produces a population $Q$ in which each variable is activated individually. By performing non-dominated sorting on $Q$ and accumulating the Pareto front ranks, the importance score of each variable is obtained. Finally, $\mathbf{cpv}$ is normalized and inversely mapped to represent the probability that each variable is nonzero. After obtaining $\mathbf{cpv}$, Algorithm~\ref{alg:init} is used to generate the initial population. It first randomly creates $N$ real solutions forming the matrix $Real$ and initializes a zero matrix $Bin$. For each individual, the number of nonzero variables is randomly determined, and a pairwise comparison is conducted to prioritize the activation of more important variables: if $cpv_m > cpv_n$, variable $x_m$ is considered more important, and $Bin_{i,m}=1$ is set; otherwise, $x_n$ is activated. The initial population $P$ is then obtained by the element-wise product $P = Real \odot Bin$.

The constructed $\mathbf{cpv}$ is embedded into Algorithm~\ref{alg:CPVSearch} to assist the first subpopulation in generating offspring solutions. Leveraging the prior information encoded in $\mathbf{cpv}$, Algorithm~\ref{alg:CPVSearch} determines whether each variable should be activated as nonzero according to its probability in $\mathbf{cpv}$, performing single-variable flips to achieve stable and efficient local exploitation. Specifically, the algorithm randomly selects two parents $\mathbf{p}$ and $\mathbf{q}$ from the parent pool $P'$ and computes the index of differing variables as $index = xor(\mathbf{bin}^p, \mathbf{bin}^q)$, which identifies the positions where the two solutions differ. Then, $\mathbf{bin}^p$ is used as the initial binary template $\mathbf{bin}^o$ for the offspring, followed by crossover and mutation operations. For crossover, one of two operations is performed with equal probability: if the random number is less than 0.5, two positions $(m,n)$ are randomly selected from the difference index, and their corresponding values in $\mathbf{cpv}$ are compared— the position with the higher $\mathbf{cpv}$ value is set to 1, prioritizing activation of variables more critical for convergence; otherwise, the position with the lower $\mathbf{cpv}$ value is set to 0, maintaining the sparsity of the solution. Similarly, for mutation, one of two operations is performed with equal probability: if the random number is less than 0.5, two positions are randomly selected from the zero set of $\mathbf{bin}^o$, and the variable with the higher $\mathbf{cpv}$ value is flipped to 1; otherwise, two positions are selected from the nonzero set, and the variable with the lower $\mathbf{cpv}$ value is set to 0. Subsequently, in the real space, the algorithm performs simulated binary crossover and polynomial mutation on the real parts of the parent solutions to produce the real offspring vector $\mathbf{real}^o$. Finally, the complete offspring solution $\mathbf{o}$ is obtained through the element-wise product of $\mathbf{bin}^o$ and $\mathbf{real}^o$, and added to the offspring population $Q$. This process is repeated until the parent pool $P'$ becomes empty, after which the full offspring population $Q$ is output.

\subsection{Annealing-Based Variable Grouping}

Before generating the offspring population through the annealing-based search strategy, the proposed variable clustering method is employed to divide all decision variables into multiple groups, each assigned with a distinct activation probability. The annealing-driven probability vector $\mathbf{apv}$ is first computed as follows:
\begin{equation}
   \label{equ:apv}
   \begin{aligned}
    \mathbf{apv}_i 
    &= \frac{1}{2}(1 - \text{rate}) 
       + \text{rate} \times 
       \frac{1}{|P|} \sum_{p \in P} \mathbf{bin}^p_i \\
    &\quad i = 1, 2, \dots, D
    \end{aligned} \ ,
\end{equation}
where $\mathbf{bin}^p_i$ denotes the binary value (0 or 1) of individual $p$ in the $i$-th dimension, $|P|$ is the population size, and $\text{rate}\in[0,1]$ represents the annealing coefficient, i.e., the ratio of consumed function evaluations to the total number of evaluations. The formula means that the proportion of nonzero values in the current population for each variable, $\frac{1}{|P|}\sum_{p\in P}\mathbf{bin}^p_i$, is linearly mapped to a dynamic interval $[(1 - \text{rate})/2,(1 + \text{rate})/2]$. In the early stages of evolution, this interval is narrow, and the probabilities of different variables concentrate around 0.5, which maintains high perturbation strength and global exploration ability. As the evolution proceeds and $\text{rate}$ increases with the number of function evaluations, the interval gradually expands, enabling a smooth transition from global exploration to local exploitation.

After obtaining $\mathbf{apv}$, Algorithm~\ref{alg:VariableClustering} adaptively determines the number of decision variables per group, $G_{\text{size}}$, by computing the average sparsity of the population, and then sorts all variable indices according to their $\mathbf{apv}$ values. Once sorted, all variables are divided sequentially into several consecutive groups, each containing $G_{\text{size}}$ variables, with the total number of groups given by $K = \lceil D / G_{\text{size}} \rceil$. After all decision variables are assigned to multiple groups, Algorithm~\ref{alg:VariableClustering} further assigns a group-level probability $p_g$ to each group, defined as the mean $\mathbf{apv}$ value of all variables within that group:
\begin{equation}
   p_g = \frac{1}{|G_g|}\sum_{i\in G_g}\mathbf{apv}_i \ .
\end{equation}

Thus, each group not only contains an independent subset of variables but also possesses a search probability that reflects its relative importance. A higher $p_g$ indicates that the variables in that group are more likely to be activated as nonzero at the current evolutionary stage. Finally, the proposed variable clustering method outputs the set of variable groups $\textit{Groups}$ and their corresponding group probabilities $\textit{Probs}$. Clearly, in the early stages of evolution, due to the high entropy of $\mathbf{apv}$, the probability distribution across groups is relatively uniform, encouraging coarse-grained global exploration. As evolution progresses and the entropy of $\mathbf{apv}$ gradually decreases, the probability differences between groups widen, allowing the algorithm to fully exploit the sparse distribution characteristics of the current population and effectively approach the sparse Pareto optimal solutions.

\subsection{Annealing Search Strategy}

\begin{algorithm}[!t]
	\caption{AnnealingSearch}
	\small
	\label{alg:APVSearch}
	\begin{algorithmic}[1]
		\REQUIRE $P^\prime$ (parent solutions), $Groups$ (variable groups), $Probs$ (probability assigned to each group)
		\ENSURE $Q$ (offspring population)
		\WHILE{$P^\prime \neq \emptyset$}
		\STATE $[\mathbf{p},\mathbf{q}] \leftarrow$ Randomly select two parents from $P^\prime$       	
		\STATE $index \leftarrow xor(\mathbf{bin}^p, \mathbf{bin}^q)$
		\STATE $P^\prime \leftarrow P^\prime \setminus \{\mathbf{p},\mathbf{q}\}$ \\
		\COMMENT{$\mathbf{bin}^p$ denotes the binary vector of solution $\mathbf{p}$}
		\STATE $\mathbf{bin}^o \leftarrow \mathbf{bin}^p$\\

		\COMMENT{Crossover}
		\STATE $g \leftarrow$ Randomly select one group index from $Groups$\;
		\STATE $G \leftarrow Groups[g]$\;
		\STATE $p_g \leftarrow Probs[g]$\;
		\IF{$rand() < p_g$}
		    \STATE Set bits in $(G \cap index)$ to 1 in $\mathbf{bin}^o$\;
		\ELSE
		    \STATE Set bits in $(G \cap index)$ to 0 in $\mathbf{bin}^o$\;
		\ENDIF\\

		\COMMENT{Mutation}
		\STATE $g^\prime \leftarrow$ Randomly select one group index from $Groups$\;
		\STATE $H \leftarrow Groups[g^\prime]$\;
        \STATE $p_{g^\prime} \leftarrow Probs[g^\prime]$\;
		\STATE $S \leftarrow$ Randomly select $\lfloor |H|/2 \rfloor$ distinct indices from $H$\;
		\IF{$rand() < p_{g^\prime}$}
		    \STATE Set all bits in $S$ to 1 in $\mathbf{bin}^o$\;
		\ELSE
		    \STATE Set all bits in $S$ to 0 in $\mathbf{bin}^o$\;
		\ENDIF\\

		\COMMENT{Crossover and mutation for real vectors}
		\STATE $\mathbf{real}^o \leftarrow$ Perform simulated binary crossover and polynomial mutation on $\mathbf{real}^p$ and $\mathbf{real}^q$
		\STATE $\mathbf{o} \leftarrow$ A solution determined by $\mathbf{bin}^o$ and $\mathbf{real}^o$
		\STATE $Q \leftarrow Q \cup \{\mathbf{o}\}$
		\ENDWHILE		
		\RETURN $Q$
	\end{algorithmic}
\end{algorithm}

Based on the above clustering results, Algorithm~\ref{alg:APVSearch} assists the second subpopulation in generating offspring solutions by directly flipping variables of entire groups according to their respective annealing probabilities. As shown in Algorithm~\ref{alg:APVSearch}, the algorithm begins by randomly selecting two parent individuals $\mathbf{p}$ and $\mathbf{q}$ from the parent pool $P'$, and computes the index of differing bits between their binary vectors as $index = xor(\mathbf{bin}^p, \mathbf{bin}^q)$. The binary vector of $\mathbf{p}$ is then copied as the offspring template $\mathbf{bin}^o$. Subsequently, the algorithm performs a group-probability-based crossover operation. It first randomly selects a group index $g$ from the set of variable groups $Groups$, and extracts the corresponding subset of variables $G$ and its associated annealing probability $p_g$. If a random number is less than $p_g$, all bits in the intersection of $G$ and $index$ are set to 1, activating key nonzero variables; otherwise, these bits are set to 0 to maintain the sparsity of the solution. Next, a mutation operation is carried out to further enhance population diversity. Similarly, another group $g'$ is randomly selected from $Groups$, and its variable subset $H$ and probability $p_{g'}$ are retrieved. From $H$, $\lfloor |H|/2 \rfloor$ variable indices are randomly chosen to form a set $S$. If a random number is less than $p_{g'}$, all variables in $S$ are set to 1; otherwise, they are set to 0. Through this mechanism, the algorithm dynamically adjusts the activation states of variable groups at different probability levels, enabling the search to cover potentially promising regions while preserving overall sparsity. After completing the binary operations, the algorithm performs simulated binary crossover and polynomial mutation on the real components $\mathbf{real}^p$ and $\mathbf{real}^q$ of the parent individuals to generate a real offspring vector $\mathbf{real}^o$. Finally, by combining the binary vector $\mathbf{bin}^o$ and the real vector $\mathbf{real}^o$ element-wise, a complete offspring individual $\mathbf{o}$ is obtained and added to the offspring set $Q$. This process is repeated until the parent pool $P'$ becomes empty.

\begin{table*}[t!]
	\renewcommand{\arraystretch}{1.1}
	\centering
	\caption{Median IGD Values of TS-SparseEA, SCEA, S-NSGA-II, MOEA/CKF, DKCA, and the Proposed PAMEA on SMOP1--SMOP8}
	\resizebox{\textwidth}{!}{
		\begin{tabular}{cccccccc}
			\toprule
			Problem&$D$&TS-SparseEA&SCEA&S-NSGA-II&MOEA/CKF&DKCA&PAMEA\\
			\midrule
			\multirow{5}{*}{SMOP1}&100&6.1981e-3 (1.91e-3) $-$&5.3545e-3 (1.81e-3) $-$&5.8031e-3 (1.94e-3) $-$&4.8698e-3 (7.96e-4) $-$&5.2734e-3 (5.60e-4) $-$&\hl{4.6504e-3 (1.28e-3)}\\
			&200&5.6513e-3 (9.40e-4) $-$&4.9855e-3 (1.02e-3) $-$&6.5459e-3 (2.48e-3) $-$&5.0500e-3 (8.86e-4) $-$&5.4861e-3 (7.32e-4) $-$&\hl{4.7024e-3 (1.08e-3)}\\
			&500&6.5372e-3 (1.40e-3) $-$&6.1696e-3 (1.17e-3) $-$&7.1686e-3 (2.13e-3) $-$&6.5962e-3 (1.79e-3) $-$&6.9876e-3 (1.79e-3) $-$&\hl{5.1463e-3 (9.91e-4)}\\
			&1000&7.9943e-3 (2.13e-3) $-$&9.0455e-3 (1.83e-3) $-$&1.1994e-2 (4.39e-3) $-$&8.3187e-3 (2.13e-3) $-$&1.0467e-2 (1.83e-3) $-$&\hl{5.0104e-3 (8.57e-4)}\\
			&5000&1.3402e-2 (1.58e-3) $-$&1.8759e-2 (1.17e-3) $-$&2.0297e-2 (4.28e-3) $-$&1.6049e-2 (4.55e-3) $-$&2.5526e-2 (1.50e-3) $-$&\hl{8.1356e-3 (1.37e-3)}\\
			\hline
			\multirow{5}{*}{SMOP2}&100&1.5041e-2 (5.61e-3) $-$&1.6384e-2 (8.33e-3) $-$&1.7160e-2 (9.29e-3) $-$&9.2340e-3 (3.14e-3) $-$&1.3276e-2 (4.24e-3) $-$&\hl{6.9847e-3 (2.71e-3)}\\
			&200&1.4955e-2 (4.90e-3) $-$&1.9238e-2 (6.32e-3) $-$&1.8861e-2 (7.34e-3) $-$&9.8892e-3 (5.44e-3) $\approx$&1.4809e-2 (4.00e-3) $-$&\hl{8.8073e-3 (6.09e-3)}\\
			&500&2.1488e-2 (5.22e-3) $-$&2.5056e-2 (5.50e-3) $-$&3.1915e-2 (1.19e-2) $-$&1.7741e-2 (6.98e-3) $-$&2.3328e-2 (4.59e-3) $-$&\hl{9.2586e-3 (4.88e-3)}\\
			&1000&2.8468e-2 (5.73e-3) $-$&3.3354e-2 (4.20e-3) $-$&4.0523e-2 (9.63e-3) $-$&3.0088e-2 (7.65e-3) $-$&3.4629e-2 (6.04e-3) $-$&\hl{8.5317e-3 (2.75e-3)}\\
			&5000&5.5916e-2 (6.71e-3) $-$&6.6440e-2 (2.79e-3) $-$&8.3305e-2 (1.83e-2) $-$&5.5968e-2 (5.05e-3) $-$&7.0693e-2 (3.55e-3) $-$&\hl{1.9548e-2 (5.45e-3)}\\
			\hline
			\multirow{5}{*}{SMOP3}&100&2.1257e-2 (3.76e-2) $-$&4.8323e-3 (7.51e-4) $-$&9.5735e-3 (6.61e-3) $-$&\hl{4.0379e-3 (1.29e-4) $\approx$}&5.1825e-3 (7.30e-4) $-$&4.1875e-3 (4.49e-4)\\
			&200&7.7957e-3 (8.79e-3) $-$&4.7093e-3 (1.00e-3) $-$&9.5223e-3 (4.54e-3) $-$&\hl{4.0654e-3 (1.91e-4) $\approx$}&5.5324e-3 (6.50e-4) $-$&4.1508e-3 (4.36e-4)\\
			&500&7.8533e-3 (2.29e-3) $-$&4.4952e-3 (6.20e-4) $\approx$&1.3027e-2 (3.90e-3) $-$&5.2986e-3 (1.34e-3) $-$&6.9424e-3 (1.68e-3) $-$&\hl{4.3092e-3 (4.90e-4)}\\
			&1000&6.7570e-3 (1.70e-3) $-$&4.4735e-3 (5.34e-4) $\approx$&1.8351e-2 (4.59e-3) $-$&1.1696e-2 (3.09e-3) $-$&9.2484e-3 (1.50e-3) $-$&\hl{4.1577e-3 (1.18e-4)}\\
			&5000&7.0603e-3 (1.38e-3) $-$&6.2459e-3 (7.46e-4) $-$&2.5149e-2 (4.92e-3) $-$&2.4789e-2 (2.02e-3) $-$&2.6127e-2 (1.85e-3) $-$&\hl{4.6441e-3 (6.58e-4)}\\
			\hline
			\multirow{5}{*}{SMOP4}&100&4.6566e-3 (2.09e-4) $-$&4.1232e-3 (6.26e-5) $\approx$&5.3516e-3 (4.80e-4) $-$&4.1303e-3 (6.71e-5) $\approx$&4.7294e-3 (1.62e-4) $-$&\hl{4.1015e-3 (5.50e-5)}\\
			&200&4.6408e-3 (1.70e-4) $-$&4.1154e-3 (5.62e-5) $\approx$&5.1903e-3 (4.74e-4) $-$&4.1382e-3 (6.00e-5) $-$&4.7050e-3 (1.88e-4) $-$&\hl{4.1081e-3 (5.00e-5)}\\
			&500&4.6004e-3 (2.02e-4) $-$&4.1243e-3 (4.78e-5) $\approx$&4.9568e-3 (3.72e-4) $-$&4.1351e-3 (7.60e-5) $\approx$&4.8345e-3 (2.61e-4) $-$&\hl{4.1177e-3 (7.95e-5)}\\
			&1000&4.5878e-3 (1.31e-4) $-$&4.1335e-3 (5.58e-5) $\approx$&4.7918e-3 (2.67e-4) $-$&4.1478e-3 (7.74e-5) $\approx$&4.7356e-3 (2.38e-4) $-$&\hl{4.1287e-3 (7.25e-5)}\\
			&5000&4.6361e-3 (1.54e-4) $-$&4.1390e-3 (5.11e-5) $\approx$&4.8585e-3 (3.20e-4) $-$&4.1644e-3 (6.17e-5) $\approx$&4.7542e-3 (2.30e-4) $-$&\hl{4.1388e-3 (6.92e-5)}\\
			\hline
			\multirow{5}{*}{SMOP5}&100&6.3279e-3 (7.03e-4) $-$&8.3525e-3 (9.35e-4) $-$&4.9449e-3 (2.91e-4) $\approx$&6.8276e-3 (3.70e-4) $-$&6.5865e-3 (4.02e-4) $-$&\hl{4.9421e-3 (2.71e-4)}\\
			&200&5.8608e-3 (5.29e-4) $-$&7.0483e-3 (7.39e-4) $-$&4.9119e-3 (1.88e-4) $-$&6.5874e-3 (3.36e-4) $-$&6.7776e-3 (4.59e-4) $-$&\hl{4.6600e-3 (1.48e-4)}\\
			&500&5.2001e-3 (2.82e-4) $-$&5.3053e-3 (2.26e-4) $-$&4.8706e-3 (2.36e-4) $-$&5.7520e-3 (2.89e-4) $-$&6.5936e-3 (3.59e-4) $-$&\hl{4.5314e-3 (1.33e-4)}\\
			&1000&5.0961e-3 (1.72e-4) $-$&4.8501e-3 (1.14e-4) $-$&4.9430e-3 (4.50e-4) $-$&5.2065e-3 (1.63e-4) $-$&6.3790e-3 (2.94e-4) $-$&\hl{4.5536e-3 (1.02e-4)}\\
			&5000&5.3993e-3 (1.75e-4) $-$&4.8049e-3 (1.07e-4) $-$&2.0949e-2 (2.37e-3) $-$&4.7595e-3 (8.12e-5) $-$&6.9241e-3 (3.36e-4) $-$&\hl{4.4916e-3 (8.95e-5)}\\
			\hline
			\multirow{5}{*}{SMOP6}&100&6.7854e-3 (7.70e-4) $-$&8.7277e-3 (1.59e-3) $-$&\hl{4.9593e-3 (2.18e-4) $+$}&6.4270e-3 (4.23e-4) $-$&8.0655e-3 (8.29e-4) $-$&6.0060e-3 (3.81e-4)\\
			&200&5.7642e-3 (3.63e-4) $-$&6.5971e-3 (8.53e-4) $-$&\hl{4.8848e-3 (2.35e-4) $+$}&5.8896e-3 (2.97e-4) $-$&8.1558e-3 (6.01e-4) $-$&5.5048e-3 (3.12e-4)\\
			&500&5.3977e-3 (2.24e-4) $\approx$&5.1649e-3 (2.22e-4) $\approx$&\hl{4.8449e-3 (2.62e-4) $+$}&5.3517e-3 (5.49e-4) $\approx$&7.4992e-3 (4.41e-4) $-$&5.2797e-3 (2.42e-4)\\
			&1000&5.2525e-3 (1.85e-4) $+$&4.8718e-3 (1.05e-4) $+$&4.8659e-3 (3.19e-4) $+$&\hl{4.6873e-3 (1.10e-4) $+$}&7.6517e-3 (3.45e-4) $-$&5.4946e-3 (2.18e-4)\\
			&5000&5.8353e-3 (2.20e-4) $-$&4.9310e-3 (7.91e-5) $+$&8.0564e-3 (3.02e-3) $-$&\hl{4.4473e-3 (4.58e-5) $+$}&8.7258e-3 (3.04e-4) $-$&5.2010e-3 (3.50e-4)\\
			\hline
			\multirow{5}{*}{SMOP7}&100&2.2862e-2 (1.13e-2) $-$&1.9697e-2 (9.91e-3) $-$&1.8785e-2 (1.18e-2) $-$&1.0401e-2 (3.80e-3) $-$&1.4581e-2 (5.24e-3) $-$&\hl{8.9460e-3 (6.13e-3)}\\
			&200&2.5382e-2 (7.27e-3) $-$&2.5342e-2 (9.32e-3) $-$&3.5302e-2 (2.58e-2) $-$&1.1059e-2 (5.17e-3) $-$&1.6524e-2 (5.15e-3) $-$&\hl{7.6315e-3 (4.02e-3)}\\
			&500&2.8032e-2 (6.87e-3) $-$&3.3958e-2 (7.53e-3) $-$&5.0097e-2 (2.29e-2) $-$&2.3888e-2 (9.57e-3) $-$&2.7769e-2 (5.85e-3) $-$&\hl{1.0385e-2 (6.48e-3)}\\
			&1000&4.0151e-2 (7.37e-3) $-$&4.7269e-2 (6.55e-3) $-$&7.4589e-2 (2.63e-2) $-$&3.7616e-2 (8.59e-3) $-$&4.3464e-2 (7.26e-3) $-$&\hl{7.0071e-3 (3.26e-3)}\\
			&5000&7.4367e-2 (8.67e-3) $-$&8.7888e-2 (4.95e-3) $-$&1.3380e-1 (2.38e-2) $-$&6.7837e-2 (9.13e-3) $-$&9.5532e-2 (4.11e-3) $-$&\hl{9.8488e-3 (2.96e-3)}\\
			\hline
			\multirow{5}{*}{SMOP8}&100&1.4378e-1 (2.77e-2) $-$&1.1566e-1 (3.36e-2) $\approx$&1.1607e-1 (2.89e-2) $\approx$&1.0907e-1 (2.40e-2) $+$&\hl{1.0138e-1 (2.59e-2) $+$}&1.3061e-1 (2.60e-2)\\
			&200&1.2532e-1 (1.98e-2) $+$&1.0510e-1 (1.92e-2) $+$&1.2653e-1 (2.08e-2) $+$&\hl{1.0457e-1 (2.15e-2) $+$}&1.1306e-1 (1.66e-2) $+$&1.4619e-1 (1.80e-2)\\
			&500&1.3287e-1 (1.85e-2) $+$&\hl{1.2014e-1 (1.52e-2) $+$}&1.4017e-1 (2.01e-2) $+$&1.2350e-1 (1.55e-2) $+$&1.3731e-1 (1.43e-2) $+$&1.7147e-1 (1.81e-2)\\
			&1000&1.3793e-1 (1.29e-2) $+$&\hl{1.2399e-1 (1.21e-2) $+$}&1.6372e-1 (2.30e-2) $+$&1.5291e-1 (1.92e-2) $+$&1.6038e-1 (1.02e-2) $+$&1.7469e-1 (1.72e-2)\\
			&5000&1.6300e-1 (6.43e-3) $+$&\hl{1.5548e-1 (5.83e-3) $+$}&2.1384e-1 (1.36e-2) $-$&2.0403e-1 (1.39e-2) $-$&2.4686e-1 (8.46e-3) $-$&1.9092e-1 (7.76e-3)\\
			\hline
			\multicolumn{2}{c}{$+/-/\approx$}&5/34/1&6/25/9&7/31/2&6/26/8&4/36/0&\\
			\bottomrule
		\end{tabular}
	}
	\label{tab:benchmark}
\end{table*}
\section{Empirical Studies}

To validate the effectiveness and efficiency of the proposed algorithm in solving various LSMOPs with different landscape characteristics, this section compares it with state-of-the-art LSMOEAs on both benchmark and real-world LSMOPs. All experiments are conducted under the same hardware and software environment to ensure consistency and reproducibility. Specifically, the experiments are performed on a computer equipped with an Intel i9-14900K processor, 96 GB of RAM, and MATLAB R2022b. To avoid performance bias caused by implementation differences, all algorithms are implemented in MATLAB and executed as well as evaluated using the multi-objective evolutionary optimization platform PlatEMO v4.13 \cite{tian2023practical}.

\subsection{Comparative Algorithms}

In this study, five state-of-the-art LSMOEAs are selected as comparison algorithms, namely TS-SparseEA~\cite{jiang2022two}, SCEA~\cite{zhang2024co}, S-NSGA-II~\cite{kropp2024improved}, MOEA/CKF~\cite{ding2024efficient}, and DKCA~\cite{li2024dynamic}. When generating real decision variables for offspring individuals, S-NSGA-II employs its specialized sparse simulated binary crossover and sparse polynomial mutation operators. In contrast, TS-SparseEA, SCEA, MOEA/CKF, DKCA, and the proposed PAMEA use the simulated binary crossover~\cite{deb1995simulated} and polynomial mutation~\cite{deb1996combined}, where the crossover probability, mutation probability, and distribution index are set to $1$, $1/D$ (with $D$ denoting the number of decision variables), and $20$, respectively. For generating binary decision variables of offspring, S-NSGA-II samples real numbers within the interval $[0, 1]$ and converts them into binary values through a rounding operation. TS-SparseEA, SCEA, MOEA/CKF, DKCA and the proposed PAMEA adopt their dedicated crossover and mutation operators as defined in the original literature. To ensure fair comparisons, all algorithms use a unified population size of $100$. Regarding termination conditions, the maximum number of function evaluations is set to $100 \times D$ for benchmark test instances and $100000$ for real-world test instances. To preserve the performance characteristics reported in the original studies, all parameter settings of the comparison algorithms strictly follow their respective original references. Each algorithm is independently executed $30$ times on each test instance to obtain statistically meaningful and stable results.

\subsection{Test Problems}

In the performance evaluation, this study employs eight benchmark LSMOPs and ten real-world LSMOPs. Specifically, SMOP1--SMOP8 are eight benchmark test problems with varying levels of difficulty, where the objective dimension $M$, the decision variable dimension $D$, and the sparsity parameter $\theta$ can all be customized as needed. Meanwhile, SR1--SR5 and PM1--PM5 correspond to sparse signal reconstruction problems and pattern mining problems based on ten datasets, respectively. In the experimental parameter settings, the objective dimension of SMOP1–SMOP8 is fixed to $M = 2$, and the decision variable dimension $D$ is set from $100$ to $5\,000$. The sparsity parameter is defined as $\theta = \lceil 0.1(D - M + 1) \rceil$ . Regarding performance metrics, the inverted generational distance (IGD)~\cite{bosman2003balance} is used to evaluate the benchmark LSMOPs. For real-world problems, since their true Pareto fronts are unknown, the hypervolume (HV)~\cite{zitzler1999multiobjective} metric is adopted to assess the quality of the solution sets obtained by each algorithm. Specifically, for IGD, 10,000 reference points are uniformly sampled from each Pareto front approximation, while for HV, the reference point is set to $(1, 1)$, since the objective values of all real-world problems considered in this study are bounded within 1. Furthermore, the Wilcoxon rank-sum test~\cite{okoye2024wilcoxon} with a significance level of 0.05 is employed to statistically analyze the experimental results, in order to determine whether significant differences exist between the proposed PAMEA and the comparison algorithms. In the results, ``+'', ``-'', and ``$\approx$'' indicate that a compared algorithm performs significantly better than, significantly worse than, or not significantly different from the proposed algorithm, respectively.

\begin{figure*}[!t]
	\centering
	\subfigure{\includegraphics[width=0.9\linewidth]{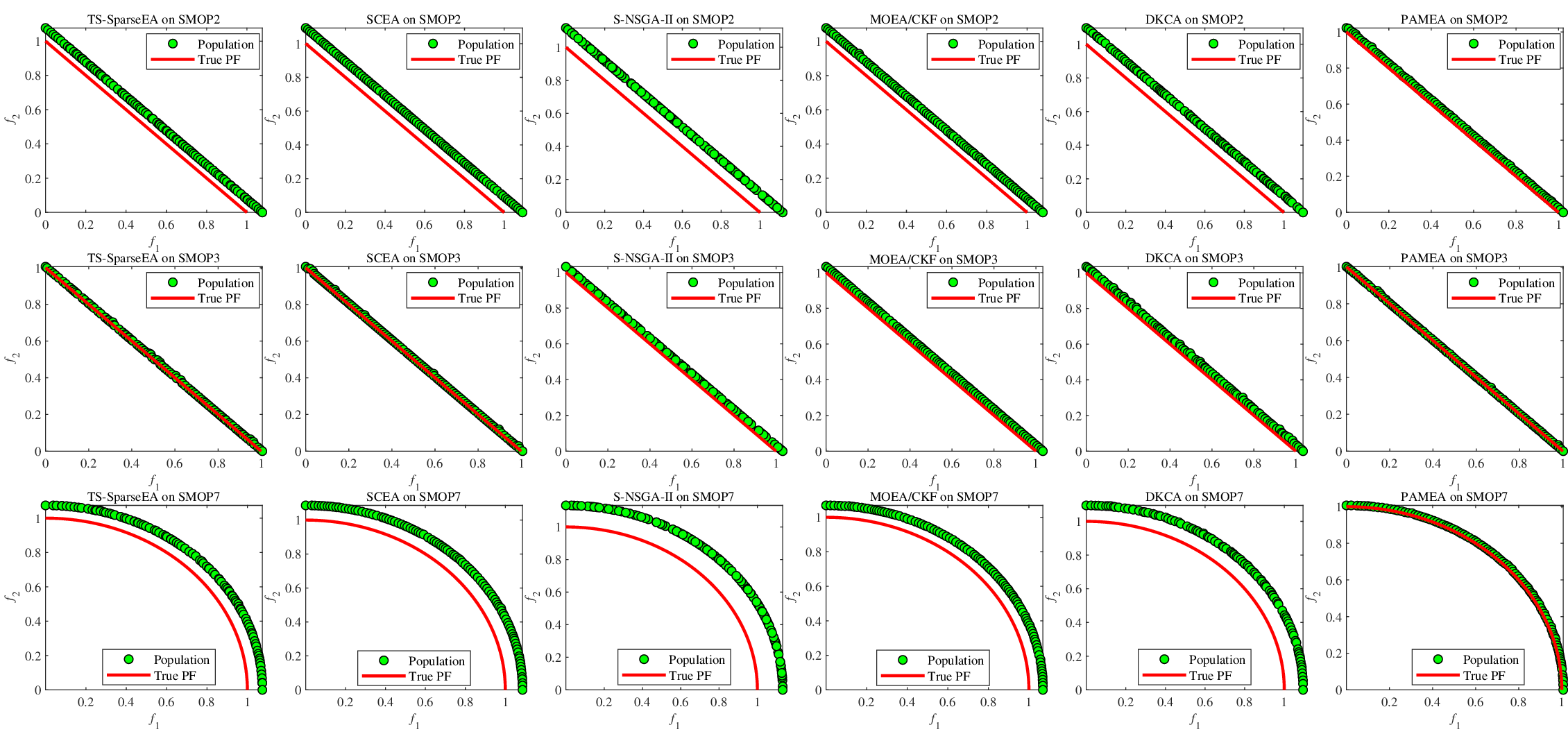}}\\
	\caption{Objectives values obtained by TS-SparseEA, SCEA, S-NSGA-II, MOEA/CKF, DKCA, and the Proposed PAMEA on SMOP2, SMOP3, and SMOP7 with 5\,000 decision variables.}
	\label{fig:smopobj}
\end{figure*}

\subsection{Results on Benchmark Problems}

Table \ref{tab:benchmark} presents the median IGD values obtained by the proposed PAMEA and the compared algorithms on SMOP1–SMOP8 over 30 independent runs. According to the statistical results, PAMEA significantly outperforms TS-SparseEA, SCEA, S-NSGA-II, MOEA/CKF, and DKCA on 34, 25, 31, 26, and 36 test instances, respectively. Therefore, the proposed PAMEA demonstrates overall superior performance compared with the existing LSMOEAs. It is worth noting that although PAMEA does not achieve the best results on certain test problems (e.g., SMOP6 and SMOP8), its IGD values remain within an acceptable range and are very close to the best results. This indicates that even though these problems may involve special variable interaction patterns that give certain algorithms advantages in specific scenarios, the proposed PAMEA still exhibits strong adaptability and stability due to its dual-entropy probability mechanism and does not show noticeable performance degradation.

\begin{figure*}[!t]
	\centering
	\subfigure{\includegraphics[width=0.86\linewidth]{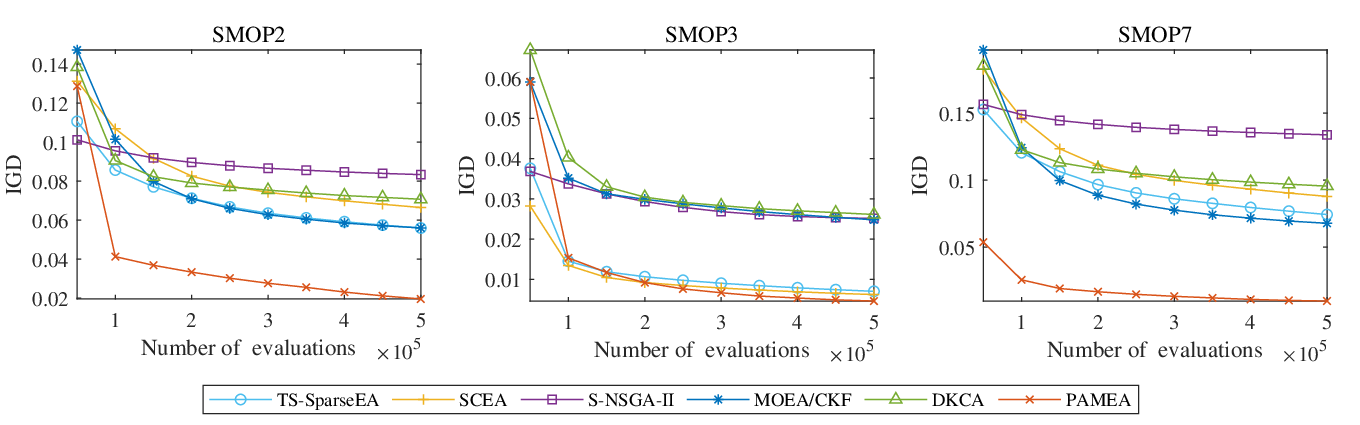}}\\
	\caption{IGD variations obtained by TS-SparseEA, SCEA, S-NSGA-II, MOEA/CKF, DKCA, and the Proposed PAMEA on SMOP2, SMOP3, and SMOP7 with 5\,000 decision variables}
	\label{fig:exp_igd}
\end{figure*}

To further examine the performance differences between the proposed algorithm and the compared algorithms, Fig. \ref{fig:smopobj} illustrates the nondominated solution sets corresponding to the median IGD values obtained by six MOEAs on SMOP2, SMOP3, and SMOP7 with 5\,000 decision variables. For SMOP3, which is characterized by deceptive features and relatively simple difficulty, the identification of nonzero variables is less challenging, and most compared algorithms are able to converge to the Pareto front. However, as the problem landscape becomes more complex, the performance gaps among algorithms become more apparent. SMOP2 exhibits both deceptive and multimodal properties, while SMOP7 is even more challenging as it not only has multimodality but also involves complex variable interactions, meaning that the decision variables are heavily coupled. Under such highly complex and strongly correlated landscapes, most compared algorithms become trapped in local optima. Although some MOEAs achieve results close to those obtained by PAMEA on the same problems, a clear performance difference still exists. To highlight this, Fig. \ref{fig:exp_igd} shows the IGD convergence curves of the six MOEAs on the same problems. It can be clearly seen that the proposed PAMEA not only achieves the best final performance but also converges significantly faster than the compared algorithms throughout the entire evolutionary process.

\begin{table*}[t!]
	\renewcommand{\arraystretch}{1.1}
	\centering
	\caption{Median HV Values of TS-SparseEA, SCEA, S-NSGA-II, MOEA/CKF, DKCA, and the Proposed PAMEA on Real-World Applications.}
	\resizebox{\textwidth}{!}
	{
		\begin{tabular}{cccccccc}
            \toprule
            Problem&$D$&TS-SparseEA&SCEA&S-NSGA-II&MOEA/CKF&DKCA&PAMEA\\
            \midrule
            \multirow{1}{*}{SR1}&256&3.6665e-1 (4.97e-3) $-$&3.6289e-1 (3.72e-3) $-$&1.3890e-1 (1.03e-2) $-$&3.6840e-1 (4.37e-3) $-$&3.8720e-1 (1.01e-3) $\approx$&\hl{3.8759e-1 (7.75e-4)}\\
            \multirow{1}{*}{SR2}&512&3.5909e-1 (7.82e-3) $-$&3.6587e-1 (1.65e-3) $-$&1.1731e-1 (7.19e-3) $-$&3.3739e-1 (6.98e-3) $-$&3.7989e-1 (2.26e-3) $\approx$&\hl{3.8040e-1 (2.58e-3)}\\
            \multirow{1}{*}{SR3}&1024&3.6442e-1 (6.77e-3) $-$&3.7955e-1 (1.13e-3) $-$&1.0965e-1 (7.31e-3) $-$&2.9762e-1 (1.17e-2) $-$&3.9638e-1 (4.38e-3) $-$&\hl{3.9858e-1 (3.11e-3)}\\
            \multirow{1}{*}{SR4}&2048&2.1233e-1 (1.70e-2) $-$&3.7946e-1 (8.06e-4) $-$&1.0145e-1 (3.38e-3) $-$&2.3633e-1 (1.49e-2) $-$&3.7638e-1 (4.28e-3) $-$&\hl{3.8138e-1 (3.08e-3)}\\
            \multirow{1}{*}{SR5}&5120&1.7211e-1 (1.50e-2) $-$&\hl{3.7478e-1 (6.80e-3) $+$}&9.5471e-2 (1.49e-3) $-$&1.6843e-1 (9.88e-3) $-$&2.9706e-1 (6.17e-3) $-$&3.5460e-1 (8.50e-3)\\
            \hline
            \multirow{1}{*}{PM1}&100&3.2881e-1 (1.85e-3) $-$&2.9139e-1 (4.58e-3) $-$&2.7645e-1 (8.73e-3) $-$&3.1044e-1 (6.79e-3) $-$&3.1779e-1 (5.15e-3) $-$&\hl{3.3413e-1 (2.19e-5)}\\
            \multirow{1}{*}{PM2}&500&1.8229e-1 (4.86e-3) $-$&1.6537e-1 (3.99e-3) $-$&1.1945e-1 (5.86e-3) $-$&1.7868e-1 (4.22e-3) $-$&1.7460e-1 (3.01e-3) $-$&\hl{1.9501e-1 (5.75e-3)}\\
            \multirow{1}{*}{PM3}&1000&2.1241e-1 (3.52e-3) $-$&1.9849e-1 (5.33e-3) $-$&1.0796e-1 (5.98e-3) $-$&2.0870e-1 (4.04e-3) $-$&2.0857e-1 (4.85e-3) $-$&\hl{2.1881e-1 (4.11e-3)}\\
            \multirow{1}{*}{PM4}&2000&1.8600e-1 (3.40e-3) $-$&1.6801e-1 (4.17e-3) $-$&1.0038e-1 (3.11e-3) $-$&1.8692e-1 (4.66e-3) $-$&1.8396e-1 (4.07e-3) $-$&\hl{1.9179e-1 (4.79e-3)}\\
            \multirow{1}{*}{PM5}&5000&1.7011e-1 (3.41e-3) $\approx$&1.5864e-1 (3.95e-3) $-$&9.7657e-2 (3.61e-3) $-$&1.6534e-1 (4.94e-3) $-$&1.6464e-1 (3.46e-3) $-$&\hl{1.7224e-1 (4.04e-3)}\\
            \hline
            \multicolumn{2}{c}{$+/-/\approx$}&0/9/1&1/9/0&0/10/0&0/10/0&0/8/2&\\
            \bottomrule
        \end{tabular}
	}
	\label{tab:realworld}
\end{table*}

\subsection{Results on Real-world Applications}

\begin{figure}[!t]
	\centering
	\subfigure{\includegraphics[width=0.78\linewidth]{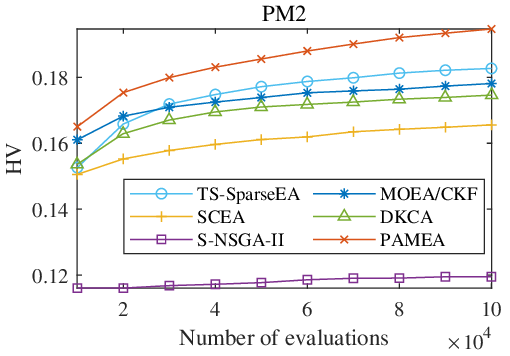}}\\
	\caption{HV variations obtained by TS-SparseEA, SCEA, S-NSGA-II, MOEA/CKF, DKCA, and the Proposed PAMEA on PM2.}
	\label{fig:exp_hv}
\end{figure}

Table \ref{tab:realworld} presents the median HV values obtained by the five compared algorithms and the proposed PAMEA on ten real-world LSMOPs. As shown in the table, PAMEA achieves significantly better HV values than the other five MOEAs and obtains the best performance on most test instances. According to the statistical results, the proposed algorithm significantly outperforms TS-SparseEA, SCEA, S-NSGA-II, MOEA/CKF, and DKCA on 9, 9, 10, 10, and 8 real-world test instances, respectively. The results in Table \ref{tab:realworld} are consistent with the trends observed in Table \ref{tab:benchmark}, further demonstrating that PAMEA not only performs well on benchmark problems but also possesses excellent optimization capability in real-world applications. To more intuitively illustrate the superiority of PAMEA, Fig. \ref{fig:exp_hv} shows the HV convergence curves of the five compared algorithms and the proposed PAMEA on PM2. It can be observed that PAMEA reaches the performance level achieved by the compared algorithms under the full evaluation budget during only the early stage of evolution. For real-world LSMOPs where function evaluations are expensive, this can significantly reduce function evaluations, thereby demonstrating the outstanding efficiency and practical advantages of PAMEA in solving real-world LSMOPs.

\subsection{Ablation Studies}

\begin{table}[t!]
	\renewcommand{\arraystretch}{1.1}
	\centering
	\caption{IGD Values Obtained by PAMEA and Its Three Variants on SMOP1--SMOP8.}
	\scalebox{0.9}{
		\begin{tabular}{cccccc}
			\toprule
			Problem&$D$&PAMEA1&PAMEA2&PAMEA3&PAMEA\\
			\midrule
			\multirow{1}{*}{SMOP1}&1000&6.9743e-3 $-$&1.2482e-2 $-$&8.8315e-3 $-$&\hl{5.0104e-3}\\
			\multirow{1}{*}{SMOP2}&1000&1.7159e-2 $-$&2.8697e-2 $-$&1.9779e-2 $-$&\hl{8.5317e-3}\\
			\multirow{1}{*}{SMOP3}&1000&1.5094e-2 $-$&1.7967e-2 $-$&4.2438e-3 $-$&\hl{4.1577e-3}\\
			\multirow{1}{*}{SMOP4}&1000&4.1512e-3 $\approx$&4.1492e-3 $\approx$&4.1374e-3 $\approx$&\hl{4.1287e-3}\\
			\multirow{1}{*}{SMOP5}&1000&4.5243e-3 $\approx$&\hl{4.4222e-3 $+$}&4.5329e-3 $\approx$&4.5536e-3\\
			\multirow{1}{*}{SMOP6}&1000&5.7049e-3 $-$&\hl{5.2524e-3 $+$}&5.6106e-3 $-$&5.4946e-3\\
			\multirow{1}{*}{SMOP7}&1000&1.3848e-2 $-$&2.0006e-2 $-$&2.2567e-2 $-$&\hl{7.0071e-3}\\
			\multirow{1}{*}{SMOP8}&1000&2.0429e-1 $-$&1.9428e-1 $-$&2.0395e-1 $-$&\hl{1.7469e-1}\\
			\hline
			\multicolumn{2}{c}{$+/-/\approx$}&0/6/2&2/5/1&0/6/2&\\
			\bottomrule
		\end{tabular}
	}
	\label{tab:abla}
\end{table}

This section conducts ablation studies to verify the effectiveness of each core component in the proposed algorithm. To minimize the interference caused by differences in other strategies, we compare the complete algorithm PAMEA with its three variants on SMOP1–SMOP8 with 1000 decision variables. Table \ref{tab:abla} reports the statistical results of PAMEA and its variants on each test instance. Specifically, PAMEA1 and PAMEA2 retain only the exploitation search strategy and the annealing search strategy, respectively, to analyze the performance differences between the two search mechanisms. PAMEA3 removes the annealing process of $\mathbf{apv}$ (i.e., the rate is always set to 1) to examine the impact of the annealing-driven adjustment of the probability vector on the final performance. The results show that, compared with the three variants, PAMEA achieves the best performance on six test instances, and the overall statistics also indicate that PAMEA outperforms the other baselines. Furthermore, in terms of statistical significance tests, PAMEA is significantly better than PAMEA1, PAMEA2, and PAMEA3 on 6, 5, and 6 test instances, respectively. These observations suggest that relying on a single search strategy often fails to simultaneously satisfy the requirements of local exploitation and global exploration, whereas coordinating exploitation search with annealing search can achieve a more effective balance between exploitation and exploration, thereby enhancing the solving capability on LSMOPs.

\section{Conclusions}

To efficiently solve LSMOPs, this paper proposes an evolutionary algorithm based on a probabilistic annealing mechanism. Specifically, the proposed algorithm constructs dual-entropy probability vectors to achieve dynamic exploration and stable exploitation in high-dimensional sparse decision spaces. The low-entropy probability vector focuses on fine-grained optimization in key regions, whereas the annealing-driven high-entropy probability vector provides strong exploratory capability in the early stage of evolution and gradually shifts toward exploitation as the annealing progresses. In addition, the proposed annealing-based variable clustering method dynamically adjusts the mapping interval of each variable group throughout the evolutionary process, thereby enhancing the algorithm’s ability to identify nonzero variables. In the experiments, a set of benchmark and real-world LSMOPs is used to evaluate the proposed algorithm, and the results show that it consistently outperforms existing LSMOEAs, delivering higher-quality solutions and faster convergence.

While the experimental results are effective for solving LSMOPs, several important directions remain to be explored. First, learning probability vectors becomes extremely challenging in super large-scale sparse spaces containing millions of decision variables. This issue may be addressed by incorporating problem-dependent dataset information to construct annealed probability vectors. Second, the dual-entropy probabilistic mechanism can be extended to estimate the likelihood of each variable being selected as a key variable for prioritized optimization, enabling the algorithm to handle LMOPs with non-sparse Pareto optimal solutions. Moreover, graph neural networks~\cite{khemani2024review} could be used to capture latent interactions among decision variables, allowing probability updates to exploit variable coupling rather than relying solely on individual contributions. Finally, integrating the learning of probability vectors with meta-learning or reinforcement learning~\cite{ma2025toward} to adaptively adjust the annealing rate and other key control parameters represents another promising direction for future work.

\ifCLASSOPTIONcaptionsoff
  \newpage
\fi
\bibliographystyle{IEEEtran}
\bibliography{references}
\end{document}